\documentclass[11pt,a4paper]{article}
\usepackage[hyperref]{emnlp2020}
\usepackage{times}
\usepackage{latexsym}
\usepackage{hyperref}
\usepackage{graphicx}
\usepackage{amsmath}
\usepackage{booktabs}
\usepackage{bm}
\usepackage{multicol}
\usepackage{svg}
\usepackage{pgfplots}
\usepackage{pgfplotstable}
\usepackage{booktabs}
\usepackage{array}
\usepackage{colortbl}
\usepackage{tikz}
\usepackage{rotating}
\pgfplotsset{compat=1.14}

\definecolor{Unbabel1}{HTML}{3843d0} 
\definecolor{Unbabel2}{HTML}{15006D} 
\definecolor{Unbabel3}{HTML}{8FA1FF}
\definecolor{Unbabel4}{HTML}{FFC466} 
\definecolor{Unbabel5}{HTML}{F9623E} 
\definecolor{Unbabel6}{HTML}{FF81A9}
\definecolor{Unbabel7}{HTML}{6ECFBD} 
\definecolor{ClassBlue}{HTML}{0492C2}

\usepackage{microtype}

\aclfinalcopy 
\def\aclpaperid{835} 

\title{{\sc Comet}: A Neural Framework for MT Evaluation}

\author{Ricardo Rei \qquad Craig Stewart \qquad Ana C Farinha \qquad Alon Lavie \\ Unbabel AI\\
  {\fontsize{10}{10}\selectfont {\texttt{\{ricardo.rei, craig.stewart, catarina.farinha, alon.lavie\}@unbabel.com}}}  
}

\date{}

\begin{document}
\maketitle
\begin{abstract}
We present {\sc Comet}, a neural framework for training multilingual machine translation evaluation models which obtains new state-of-the-art levels of correlation with human judgements. Our framework leverages recent breakthroughs in cross-lingual pretrained language modeling resulting in highly multilingual and adaptable MT evaluation models that exploit information from both the source input and a target-language reference translation in order to more accurately predict MT quality. To showcase our framework, we train three models with different types of human judgements: \textit{Direct Assessments}, \textit{Human-mediated Translation Edit Rate} and \textit{Multidimensional Quality Metrics}. Our models achieve new state-of-the-art performance on the WMT 2019 Metrics shared task and demonstrate robustness to high-performing systems. 

\end{abstract}

\section{Introduction}

Historically, metrics for evaluating the quality of machine translation (MT) have relied on assessing the similarity between an MT-generated hypothesis and a human-generated reference translation in the target language.  Traditional metrics have focused on basic, lexical-level features such as counting the number of matching n-grams between the MT hypothesis and the reference translation. Metrics such as {\sc Bleu} \cite{papineni-etal-2002-bleu} and {\sc Meteor} \cite{banerjee-lavie-meteor2009} remain popular as a means of evaluating MT systems due to their light-weight and fast computation. 

Modern neural approaches to MT result in much
higher quality of translation that often deviates
from monotonic lexical transfer between languages. 
For this reason, it has become increasingly evident that we can no longer rely on metrics such as {\sc Bleu} to provide an accurate estimate of the quality of MT \cite{barrault-etal-2019-findings}.

While an increased research interest in neural methods for training MT models and systems has resulted in a recent, dramatic improvement in MT quality, MT evaluation has fallen behind. The MT research community still relies largely on outdated metrics and no new, widely-adopted standard has emerged.  In 2019, the WMT News Translation Shared Task received a total of 153 MT system submissions \cite{barrault-etal-2019-findings}. The Metrics Shared Task of the same year saw only 24 submissions, almost half of which were entrants to the Quality Estimation Shared Task, adapted as metrics \cite{ma-etal-2019-results}. 

The findings of the above-mentioned task highlight two major challenges to MT evaluation which we seek to address herein \cite{ma-etal-2019-results}. Namely, that current metrics \textbf{struggle to accurately correlate with human judgement at segment level} and \textbf{fail to adequately differentiate the highest performing MT systems}.


In this paper, we present {\sc Comet}\footnote{\textbf{C}rosslingual \textbf{
O}ptimized \textbf{M}etric for \textbf{E}valuation of \textbf{T}ranslation.}, a PyTorch-based framework for training highly multilingual and adaptable MT evaluation models that can function as metrics. Our framework takes advantage of recent breakthroughs in cross-lingual language modeling \cite{laser2019-Artetxe, devlin-etal-2019-bert, NIPS2019_8928, conneau2019unsupervised} to generate prediction estimates of human judgments such as \textit{Direct Assessments} (DA) \cite{graham-etal-2013-continuous}, \textit{Human-mediated Translation Edit Rate} (HTER) \cite{Snover06astudy} and metrics compliant with the \textit{Multidimensional Quality Metric} framework \cite{mqm}. 

Inspired by recent work on Quality Estimation (QE) that demonstrated that it is possible to achieve high levels of correlation with human judgements even without a reference translation  \cite{fonseca-etal-2019-findings}, we propose a novel approach for incorporating the source-language input into our MT evaluation models. Traditionally only QE models have made use of the source input, whereas MT evaluation metrics rely instead on the reference translation. As in \cite{takahashi-etal-2020-automatic}, we show that using a multilingual embedding space allows us to leverage information from all three inputs and demonstrate the value added by the source as input to our MT evaluation models.

To illustrate the effectiveness and flexibility of the {\sc Comet} framework, we train three models that estimate different types of human judgements and show promising progress towards both better correlation at segment level and robustness to high-quality MT. 


We will release both the {\sc Comet} framework and the trained MT evaluation models described in this paper to the research community upon publication.

\section{Model Architectures}
\label{sec:model}
Human judgements of MT quality usually come in the form of segment-level scores, such as DA, MQM and HTER. For DA, it is common practice to convert scores into relative rankings ({\small DA}RR) when the number of annotations per segment is limited  \cite{bojar-etal-2017-results, ma-etal-2018-results, ma-etal-2019-results}. This means that, for two MT hypotheses $h_i$ and $h_j$ of the same source $s$, if the DA score assigned to $h_i$ is higher than the score assigned to $h_j$, $h_i$ is regarded as a ``better'' hypothesis.\footnote{In the WMT Metrics Shared Task, if the difference between the DA scores is not higher than 25 points, those segments are excluded from the {\scriptsize DA}RR data.} To encompass these differences, our framework supports two distinct architectures: The {\bf Estimator model} and the {\bf Translation Ranking model}. The fundamental difference between them is the training objective. While the Estimator is trained to regress directly on a quality score, the Translation Ranking model is trained to minimize the distance between a ``better'' hypothesis and both its corresponding reference and its original source. Both models are composed of a cross-lingual encoder and a pooling layer.

\subsection{Cross-lingual Encoder}
\label{ssec:encoder}
The primary building block of all the models in our framework is a pretrained, cross-lingual model such as multilingual BERT \cite{devlin-etal-2019-bert}, XLM \cite{NIPS2019_8928} or XLM-RoBERTa \cite{conneau2019unsupervised}. These models contain several transformer encoder layers that are trained to reconstruct masked tokens by uncovering the relationship between those tokens and the surrounding ones. When trained with data from multiple languages this pretrained objective has been found to be highly effective in cross-lingual tasks such as document classification and natural language inference \cite{conneau2019unsupervised}, generalizing well to unseen languages and scripts \citep{pires-etal-2019-multilingual}. For the experiments in this paper, we rely on XLM-RoBERTa (base) as our encoder model.

Given an input sequence 
$x = \left[x_0, x_1, ..., x_n\right]$, the encoder
produces an embedding
$\bm{e}_{j}^{(\ell)}$ for each token $x_j$ and each layer $\ell \in \{0,1,...,k\}$. In our framework, we apply this process to the source, MT hypothesis, and reference in order to map them into a shared feature space.

\subsection{Pooling Layer}
\label{ssec:pooling}
The embeddings generated by the last layer of the pretrained encoders are usually used for fine-tuning models to new tasks. However, \citep{tenney-etal-2019-bert} showed that different layers within the network can capture linguistic information that is relevant for different downstream tasks. In the case of MT evaluation, \citep{ZhangBERTScore} showed that different layers can achieve different levels of correlation and that utilizing only the last layer often results in inferior performance. In this work, we used the approach described in \citet{peters-etal-2018-deep} and pool information from the most important encoder layers into a single embedding for each token, $\bm{e}_{j}$, by using a layer-wise attention mechanism. This embedding is then computed as: 
\vspace{-5pt}
\begin{equation}
\label{eq:attention}
    \bm{e}_{x_j} = \mu {E}_{x_j}^\top \bm{\alpha}
\end{equation}
where $\mu$ is a trainable weight coefficient, $\bm{E}_{j} = [\bm{e}_{j}^{(0)}, \bm{e}_{j}^{(1)}, \dots\, \bm{e}_{j}^{(k)}]$ corresponds to the vector of layer embeddings for token $x_j$, and $\bm{\alpha} = \textrm{softmax} ([\alpha^{(1)}, \alpha^{(2)}, \dots, \alpha^{(k)}])$ is a vector corresponding to the layer-wise trainable weights. In order to avoid overfitting to the information contained in any single layer, we used layer dropout \cite{kondratyuk-straka-2019-75}, in which with a probability $p$ the weight $\alpha^{(i)}$ is set to $-\infty$.

Finally, as in \cite{reimers-gurevych-2019-sentence}, we apply average pooling to the resulting word embeddings to derive a sentence embedding for each segment.

\subsection{Estimator Model}
\label{ssec:estimator}


\begin{figure*}
    \centering
    \begin{minipage}{.48\textwidth}
        \centering
        \includegraphics[width=1.0\linewidth]{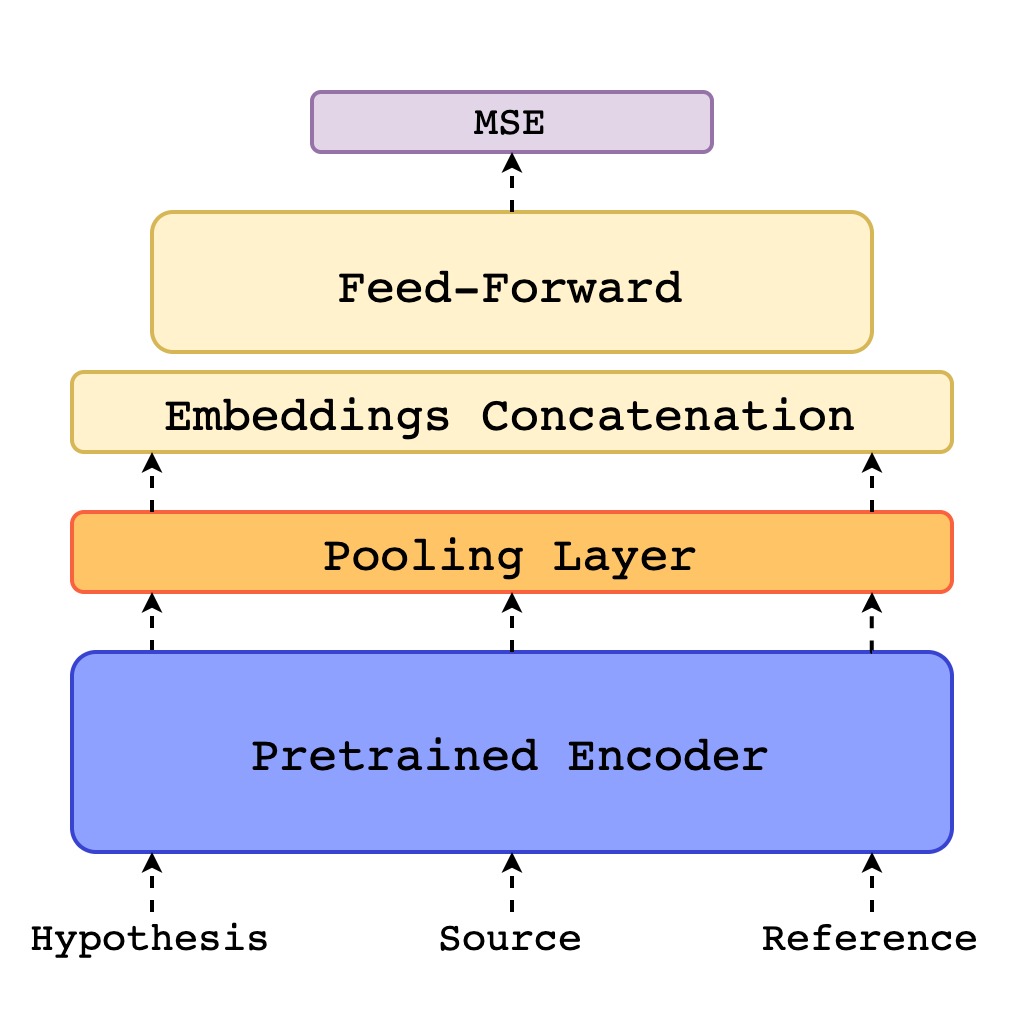}
        \caption{Estimator model architecture. The source, hypothesis and reference are independently encoded using a pretrained cross-lingual encoder. The resulting word embeddings are then passed through a pooling layer to create a sentence embedding for each segment. Finally, the resulting sentence embeddings are combined and concatenated into one single vector that is passed to a feed-forward regressor. The entire model is trained by minimizing the Mean Squared Error (MSE). }
        \label{fig:estimator}
    \end{minipage}%
    \hfill
    \begin{minipage}{.48\textwidth}
        \centering
        \vspace{25pt}
        \includegraphics[width=1.0\linewidth]{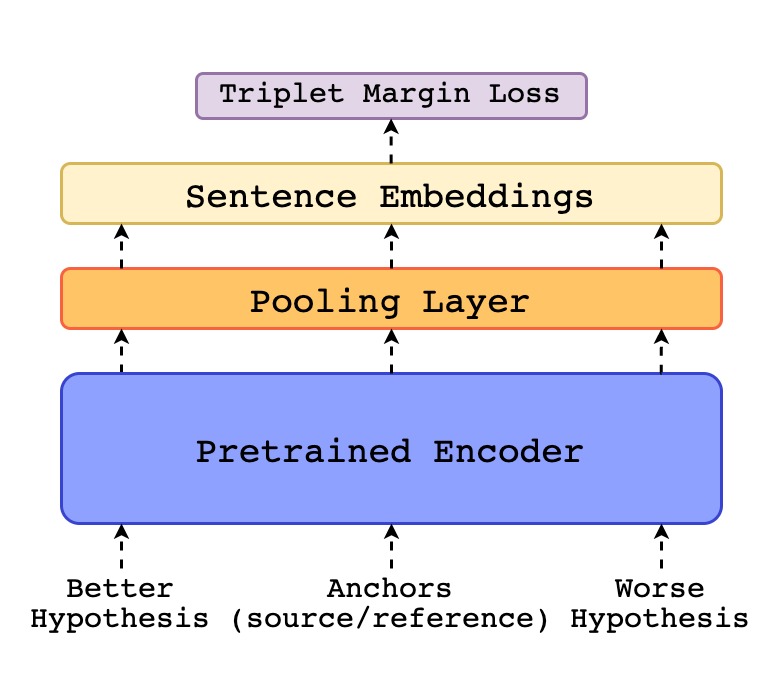}
        \caption{Translation Ranking model architecture. This architecture receives 4 segments: the source, the reference, a ``better'' hypothesis, and a ``worse'' one. These segments are independently encoded using a pretrained cross-lingual encoder and a pooling layer on top. Finally, using the triplet margin loss \cite{SchroffKP15} we optimize the resulting embedding space to minimize the distance between the ``better'' hypothesis and the ``anchors'' (source and reference).}
        \label{fig:ranking_model}
    \end{minipage}
\end{figure*}

Given a $d$-dimensional sentence embedding for the source, the hypothesis, and the reference, we adopt the approach proposed in RUSE \cite{shimanaka-etal-2018-ruse} and extract the following combined features:

\begin{itemize}
    \item Element-wise source product: $\bm{h}  \odot  \bm{s}$
    \item Element-wise reference product: $\bm{h}  \odot  \bm{r}$
    \item Absolute element-wise source difference: $|\bm{h} - \bm{s}|$
    \item Absolute element-wise reference difference: $|\bm{h} - \bm{r}|$
\end{itemize}

These combined features are then concatenated to the reference embedding $\bm{r}$ and hypothesis embedding $\bm{h}$ into a single vector $\bm{x} = [\bm{h}; \bm{r}; \bm{h}  \odot  \bm{s}; \bm{h} \odot \bm{r}; |\bm{h} - \bm{s}|; |\bm{h} - \bm{r}|]$ that serves as input to a feed-forward regressor. The strength of these features is in highlighting the differences between embeddings in the semantic feature space. 

The  model is then trained to minimize the mean squared error between the predicted scores and quality assessments (DA, HTER or MQM). Figure~\ref{fig:estimator} illustrates the proposed architecture. 

Note that we chose not to include the raw source embedding ($\bm{s}$) in our concatenated input. Early experimentation revealed that the value added by the source embedding as extra input features to our regressor was negligible at best. A variation on our HTER estimator model trained with the vector $\bm{x} = [\bm{h}; \bm{s}; \bm{r}; \bm{h}  \odot  \bm{s}; \bm{h} \odot \bm{r}; |\bm{h} - \bm{s}|; |\bm{h} - \bm{r}|]$ as input to the feed-forward only succeed in boosting segment-level performance in 8 of the 18 language pairs outlined in section \ref{sec:results} below and the average improvement in Kendall's Tau in those settings was +0.0009. As noted in \citet{zhao2020limitations}, while cross-lingual pretrained models are adaptive to multiple languages, the feature space between languages is poorly aligned. On this basis we decided in favor of excluding the source embedding on the intuition that the most important information comes from the reference embedding and reducing the feature space would allow the model to focus more on relevant information. This does not however negate the general value of the source to our model; where we include combination features such as $\bm{h} \odot \bm{s}$ and $|\bm{h} - \bm{s}|$ we do note gains in correlation as explored further in section \ref{sect:source-value} below.


\subsection{Translation Ranking Model}
\label{ssec:ranking}

Our Translation Ranking model (Figure \ref{fig:ranking_model}) receives as input a tuple $\chi = (s, h^{+}, h^{-}, r)$ where $h^{+}$ denotes an hypothesis that was ranked higher than another hypothesis $h^{-}$. We then pass $\chi$ through our cross-lingual encoder and pooling layer to obtain a sentence embedding for each segment in the $\chi$. Finally, using the embeddings $\{\bm{s}, \bm{h^{+}}, \bm{h^{-}}, \bm{r}\}$, we compute the triplet margin loss \cite{SchroffKP15} in relation to the source and reference:
\vspace{-10pt}
\begin{equation}
\label{eq:tloss}
\begin{split}
L(\chi)  &= L(\bm{s}, \bm{h^{+}}, \bm{h^{-}}) + L(\bm{r}, \bm{h^{+}}, \bm{h^{-}})
\end{split}
\end{equation}
where: \begin{equation}
\label{eq:tloss1}
\begin{split}
L(\bm{s}, \bm{h^{+}}, \bm{h^{-}}) = \\
\max\{0, d(&\bm{s}, \bm{h^{+}})\ - d(\bm{s}, \bm{h^{-}}) + \epsilon\}
\end{split}
\end{equation}

\begin{equation}
\label{eq:tloss2}
\begin{split}
L(\bm{r}, \bm{h^{+}}, \bm{h^{-}}) = \\
\max\{0, d(&\bm{r}, \bm{h^{+}})\ - d(\bm{r}, \bm{h^{-}}) + \epsilon\}
\end{split}
\end{equation}
$d(\bm{u}, \bm{v})$ denotes the euclidean distance between $\bm{u}$ and $\bm{v}$ and $\epsilon$ is a margin. Thus, during training the model optimizes the embedding space so the distance between the anchors ($\bm{s}$ and $\bm{r}$) and the ``worse'' hypothesis $\bm{h^{-}}$ is greater by at least $\epsilon$ than the distance between the anchors and ``better'' hypothesis $\bm{h^{+}}$.

During inference, the described model receives a triplet $(s, \hat{h}, r)$ with only one hypothesis. The quality score assigned to $\hat{h}$ is the harmonic mean between the distance to the source $d(\bm{s}, \bm{\hat{h}})$ and the distance to the reference $d(\bm{r}, \bm{\hat{h}})$:
\begin{equation}
\label{eq:ranking_inference}
\begin{split}
f(s, \hat{h}, r) = \frac{2 \times d(\bm{r}, \bm{\hat{h}}) \times d(\bm{s}, \bm{\hat{h}})}{d(\bm{r}, \bm{\hat{h}}) + d(\bm{s}, \bm{\hat{h}})}
\end{split}
\end{equation}

Finally, we convert the resulting distance into a similarity score bounded between 0 and 1 as follows:
\begin{equation}
\label{eq:ranking_similarity}
\begin{split}
\hat{f}(s, \hat{h}, r) = \frac{1}{1+f(s, \hat{h}, r)}
\end{split}
\end{equation}

\section{Corpora}
\label{sec:Corpora}
To demonstrate the effectiveness of our described model architectures (section \ref{sec:model}), we train three MT evaluation models where each model targets a different type of human judgment.  To train these models, we use data from three different corpora: the QT21 corpus, the {\small DA}RR from the WMT Metrics shared task (2017 to 2019) and a proprietary MQM annotated corpus. 

\subsection{The QT21 corpus}
\label{sec:qt21}
The QT21 corpus is a publicly available\footnote{QT21 data: \url{https://lindat.mff.cuni.cz/repository/xmlui/handle/11372/LRT-2390}} dataset containing industry generated sentences from either an information technology or life sciences domains \cite{specia-etal_MTSummit:2017}. This corpus contains a total of 173K tuples with source sentence, respective human-generated reference, MT hypothesis (either from a phrase-based statistical MT or from a neural MT), and post-edited MT (PE). 
The language pairs represented in this corpus are: English to German (en-de), Latvian (en-lt) and Czech (en-cs), and German to English (de-en). 

The HTER score is obtained by computing the translation edit rate (TER) \cite{Snover06astudy} between the MT hypothesis and the corresponding PE. 
Finally, after computing the HTER for each MT, we built a training dataset $D = \{s_i, h_i, r_i, y_i\}_{n=1}^N$, where $s_i$ denotes the source text, $h_i$ denotes the MT hypothesis, $r_i$ the reference translation, and $y_i$ the HTER score for the hypothesis $h_i$. In this manner we seek to learn a regression $f(s, h, r) \rightarrow y $ that predicts the human-effort required to correct the hypothesis by looking at the source, hypothesis, and reference (but not the post-edited hypothesis).

\subsection{The WMT {\small DA}RR corpus}
\label{sec:daRR}

Since 2017, the organizers of the WMT News Translation Shared Task \cite{barrault-etal-2019-findings} have collected human judgements in the form of adequacy DAs \cite{graham-etal-2013-continuous, graham-etal-2014-machine, graham_baldwin_moffat_zobel_2017}. These DAs are then mapped into relative rankings ({\small DA}RR) \cite{ma-etal-2019-results}. The resulting data for each year (2017-19) form a dataset $D = \{s_i, h_i^+, h_i^-, r_i\}_{n=1}^N$ where $h_i^+$ denotes a ``better'' hypothesis and $h_i^-$ denotes a ``worse'' one.
Here we seek to learn a function $r(s, h, r)$ such that the score assigned to $h_i^+$ is strictly higher than the score assigned to $h_i^-$ ($r(s_i, h_i^+, r_i) > r(s_i, h_i^-, r_i)$). This data\footnote{The raw data for each year of the WMT Metrics shared task is publicly available in the results page (2019 example: \url{http://www.statmt.org/wmt19/results.html}). Note, however, that in the \texttt{README} files it is highlighted that this data is not well documented and the scripts occasionally require custom utilities that are not available.} contains a total of 24 high and low-resource language pairs such as Chinese to English (zh-en) and English to Gujarati (en-gu).

\subsection{The MQM corpus}
\label{sec:mqm}

The MQM corpus is a proprietary internal database of MT-generated translations of customer support chat messages that were annotated according to the guidelines set out in \citet{mqm_guidelines}. This data contains a total of 12K tuples, covering 12 language pairs from English to: German (en-de), Spanish (en-es), Latin-American Spanish (en-es-latam), French (en-fr), Italian (en-it), Japanese (en-ja), Dutch (en-nl), Portuguese (en-pt), Brazilian Portuguese (en-pt-br), Russian (en-ru), Swedish (en-sv), and Turkish (en-tr). Note that in this corpus English is always seen as the source language, but never as the target language. 
Each tuple consists of a source sentence, a human-generated reference, a MT hypothesis, and its MQM score, derived from error annotations by one (or more) trained annotators. The MQM metric referred to throughout this paper is an internal metric defined in accordance with the MQM framework \citep{mqm} (MQM). Errors are annotated under an internal typology defined under three main error types; `Style', `Fluency' and `Accuracy'. Our MQM scores range from $-\infty$ to 100 and are defined as:

\begin{equation}
\label{eq:mqm}
\begin{split}
\text{\footnotesize MQM} = 100 - \frac{{I}_{\text{\scriptsize Minor}} + 5 \times {I}_{\text{\scriptsize Major}} 
+ 10 \times {I}_{\text{\scriptsize Crit.}}}{\text{\footnotesize Sentence Length} \times 100}
\end{split}
\end{equation}
where ${I}_{\text{\scriptsize Minor}}$ denotes the number of minor errors, ${I}_{\text{\scriptsize Major}}$ the number of major errors and ${I}_{\text{\scriptsize Crit.}}$ the number of critical errors. 

Our MQM metric takes into account the severity of the errors identified in the MT hypothesis, leading to a more fine-grained metric than HTER or DA. When used in our experiments, these values were divided by 100 and truncated at 0. As in section~\ref{sec:qt21}, we constructed a training dataset $D = \{s_i, h_i, r_i, y_i\}_{n=1}^N$, where $s_i$ denotes the source text, $h_i$ denotes the MT hypothesis, $r_i$ the reference translation, and $y_i$ the MQM score for the hypothesis $h_i$.

\begin{table*}[!ht]
\centering
\caption{Kendall's Tau ($\tau$) correlations on language pairs with English as source for the WMT19 Metrics {\footnotesize DA}RR corpus. For {\sc Bertscore} we report results with the default encoder model for a complete comparison, but also with XLM-RoBERTa (base) for fairness with our models. The values reported for YiSi-1 are taken directly from the shared task paper \cite{ma-etal-2019-results}.}
\label{tab:english-to-x2019}
\begin{tabular}{lcccccccc}
\hline
\textbf{Metric}             & \textbf{en-cs} & \textbf{en-de} & \textbf{en-fi} & \textbf{en-gu} & \textbf{en-kk} & \textbf{en-lt} & \textbf{en-ru} & \textbf{en-zh} \\ \specialrule{1.5pt}{1pt}{1pt}
{\sc Bleu}                        & 0.364          & 0.248          & 0.395          & 0.463          & 0.363          & 0.333          & 0.469          & 0.235          \\
{\sc chrF}                        & 0.444          & 0.321          & 0.518          & 0.548          & 0.510          & 0.438          & 0.548          & 0.241          \\
{\sc YiSi-1}              & 0.475  &   0.351 &   0.537 & 0.551 & 0.546 & 0.470 & 0.585 & 0.355 \\
{\sc Bertscore} {\footnotesize ({default})}          & 0.500          & 0.363          & 0.527          & 0.568          & 0.540          & 0.464          & 0.585          & 0.356          \\
{\sc Bertscore} {\footnotesize ({xlmr-base})} & 0.503          & 0.369          & 0.553          & 0.584          & 0.536          & 0.514          & 0.599          & 0.317          \\ \hline
{\sc Comet-hter}              & 0.524          & 0.383          & 0.560          & 0.552          & 0.508          & 0.577          & 0.539          & 0.380          \\
{\sc Comet-mqm}               & 0.537          & 0.398          & 0.567          & 0.564          & 0.534          & 0.574          & \textbf{0.615}          & 0.378          \\
{\sc Comet-rank}                 & \textbf{0.603} & \textbf{0.427} & \textbf{0.664} & \textbf{0.611} & \textbf{0.693} & \textbf{0.665} & 0.580 & \textbf{0.449} \\ \hline
\end{tabular}
\end{table*}

\section{Experiments}
\label{sec:experiments}
We train two versions of the Estimator model described in section \ref{ssec:estimator}: one that regresses on HTER (\textbf{{\sc Comet-hter}}) trained with the QT21 corpus, and another that regresses on our proprietary implementation of MQM (\textbf{{\sc Comet-mqm}}) trained with our internal MQM corpus. For the Translation Ranking model, described in section \ref{ssec:ranking}, we train with the WMT {\small DA}RR corpus from 2017 and 2018 (\textbf{{\sc Comet-rank}}). In this section, we introduce the training setup for these models and corresponding evaluation setup.

\subsection{Training Setup}
\label{ssec:estimator_setup}
The two versions of the Estimators (\textbf{{\sc Comet-HTER/MQM}}) share the same training setup and hyper-parameters (details are included in the Appendices). For training, we load the pretrained encoder and initialize both the pooling layer and the feed-forward regressor. Whereas the layer-wise scalars $\bm{\alpha}$ from the pooling layer are initially set to zero, the weights from the feed-forward are initialized randomly. During training, we divide the model parameters into two groups: the encoder parameters, that include the encoder model and the scalars from $\bm{\alpha}$; and the regressor parameters, that include the parameters from the top feed-forward network. We apply gradual unfreezing and discriminative learning rates \cite{howard-ruder-2018-universal}, meaning that the encoder model is frozen for one epoch while the feed-forward is optimized with a learning rate of $3\mathrm{e}{-5}$. After the first epoch, the entire model is fine-tuned but the learning rate for the encoder parameters is set to $1\mathrm{e}{-5}$ in order to avoid catastrophic forgetting.

In contrast with the two Estimators, for the \textbf{{\sc Comet-rank}} model we fine-tune from the outset. Furthermore, since this model does not add any new parameters on top of XLM-RoBERTa (base) other than the layer scalars $\bm{\alpha}$, we use one single learning rate of $1\mathrm{e}{-5}$ for the entire model.

\subsection{Evaluation Setup}
\label{ssec:metrics}
We use the test data and setup of the WMT 2019 Metrics Shared Task \cite{ma-etal-2019-results} in order to compare the {\sc Comet} models with the top performing submissions of the shared task and other recent state-of-the-art metrics such as {\sc Bertscore} and {\sc Bleurt}.\footnote{To ease future research we will also provide, within our framework, detailed instructions and scripts to run other metrics such as {\sc chrF}, {\sc Bleu}, {\sc Bertscore}, and {\sc Bleurt}} The evaluation method used is the official Kendall's Tau-like formulation, $\tau$, from the WMT 2019 Metrics Shared Task \cite{ma-etal-2019-results} defined as:

\begin{equation}
    \label{eq:kendall}
    \tau = \frac{\textit{Concordant} - \textit{Discordant}}{\textit{Concordant} + \textit{Discordant}}
\end{equation}
where \textit{Concordant} is the number of times a metric assigns a higher score to the ``better'' hypothesis $h^+$ and \textit{Discordant} is the number of times a metric assigns a higher score to the ``worse'' hypothesis $h^-$ or the scores assigned to both hypotheses is the same. 

As mentioned in the findings of \cite{ma-etal-2019-results},  segment-level correlations of all submitted metrics were frustratingly low. Furthermore, all submitted metrics exhibited a dramatic lack of ability to correctly rank strong MT systems.  To evaluate whether our new MT evaluation models better address this issue, we followed the described evaluation setup used in the analysis presented in \cite{ma-etal-2019-results}, where correlation levels are examined for portions of the {\small DA}RR data that include only the top 10, 8, 6 and 4 MT systems. 

\begin{table*}[!ht]
\centering
\caption{Kendall's Tau ($\tau$) correlations on language pairs with English as a target for the WMT19 Metrics {\footnotesize DA}RR corpus. As for {\sc Bertscore}, for {\sc Bleurt} we report results for two models: the base model, which is comparable in size with the encoder we used and the large model that is twice the size.}
\label{tab:x-to-english2019}
\begin{tabular}{llllllll}
\hline
\textbf{Metric}              & \textbf{de-en} & \textbf{fi-en} & \textbf{gu-en} & \textbf{kk-en} & \textbf{lt-en} & \textbf{ru-en} & \textbf{zh-en} \\ \specialrule{1.5pt}{1pt}{1pt}
{\sc Bleu}                        & 0.053          & 0.236          & 0.194          & 0.276          & 0.249          & 0.177          & 0.321          \\
{\sc chrF}                          & 0.123          & 0.292          & 0.240          & 0.323          & 0.304          & 0.115          & 0.371          \\
{\sc YiSi-1}   & 0.164     & 0.347       & 0.312       & \textbf{0.440}  & 0.376       & 0.217     & 0.426       \\
{\sc Bertscore} {\footnotesize ({default})}       & 0.190          & 0.354          & 0.292          & 0.351          & 0.381          & \textbf{0.221}          & 0.432          \\
{\sc Bertscore} {\footnotesize ({xlmr-base})} & 0.171          & 0.335          & 0.295          & 0.354          & 0.356          & 0.202          & 0.412          \\
{\sc Bleurt} {\footnotesize ({base-128})}    & 0.171          & 0.372          & 0.302          & 0.383     & 0.387          & 0.218          & 0.417          \\
{\sc Bleurt} {\footnotesize ({large-512})}            & 0.174          & 0.374          & 0.313          & 0.372          & 0.388          & 0.220          & 0.436    \\ \hline
{\sc Comet-hter}               & 0.185          & 0.333          & 0.274          & 0.297          & 0.364          & 0.163          & 0.391          \\
{\sc Comet-mqm}                & \textbf{0.207}     & 0.343          & 0.282          & 0.339          & 0.368          & 0.187          & 0.422          \\
{\sc Comet-rank}  & 0.202  & \textbf{0.399}    & \textbf{0.341}    & 0.358          & \textbf{0.407}         & 0.180 & \textbf{0.445}          \\ \hline
\end{tabular}
\end{table*}

\section{Results}
\label{sec:results}

\subsection{From English into X}
\label{ssec:from-English}
Table \ref{tab:english-to-x2019} shows results for all eight language pairs with English as source.  We contrast our three {\sc Comet} models against baseline metrics such as {\sc Bleu} and {\sc chrF}, the 2019 task winning metric {\sc YiSi-1}, as well as the more recent {\sc Bertscore}.  We observe that across the board our three models trained with the {\sc Comet} framework outperform, often by significant margins, all other metrics.  Our {\small DA}RR Ranker model outperforms the two Estimators in seven out of eight language pairs. Also, even though the MQM Estimator is trained on only 12K annotated segments, it performs roughly on par with the HTER Estimator for most language-pairs, and outperforms all the other metrics in en-ru. 

\subsection{From X into English}
\label{sect:into-english}
Table \ref{tab:x-to-english2019} shows results for the seven to-English language pairs. Again, we contrast our three {\sc Comet} models against baseline metrics such as {\sc Bleu} and {\sc chrF}, the 2019 task winning metric {\sc YiSi-1}, as well as the recently published metrics {\sc Bertscore} and {\sc Bleurt}. As in Table \ref{tab:english-to-x2019} the {\small DA}RR model shows strong correlations with human judgements outperforming the recently proposed English-specific {\sc Bleurt} metric in five out of seven language pairs. Again, the MQM Estimator shows surprising strong results despite the fact that this model was trained with data that did not include English as a target. Although the encoder used in our trained models is highly multilingual, we hypothesise that this powerful ``zero-shot'' result is due to the inclusion of the source in our models.

\subsection{Language pairs not involving English}
\label{sect:no-english}
\begin{table}[!ht]
\centering
\caption{Kendall's Tau ($\tau$) correlations on language pairs not involving English for the WMT19 Metrics {\small DA}RR corpus.}
\label{tab:not-english2019}
\begin{tabular}{llll}
\hline
\textbf{Metric}              & \textbf{de-cs} & \textbf{de-fr} & \textbf{fr-de} \\ \specialrule{1.5pt}{1pt}{1pt}
{\sc Bleu}                         & 0.222          & 0.226          & 0.173          \\
{\sc chrF}                         & 0.341          & 0.287          & 0.274          \\
{\sc YiSi-1}              & 0.376 & 0.349 & 0.310 \\
{\sc Bertscore} {\footnotesize ({default})}          & 0.358          & 0.329          & 0.300          \\
{\sc Bertscore} {\footnotesize ({xlmr-base})} & 0.386    & 0.336          & 0.309          \\ \hline
{\sc Comet-hter}               & 0.358          & 0.397          & 0.315          \\
{\sc Comet-mqm}                & 0.386          & 0.367          & 0.296          \\
{\sc Comet-rank}                  &  \textbf{0.389}          & \textbf{0.444}          & \textbf{0.331}          \\ \hline
\end{tabular}
\end{table}

All three of our {\sc Comet} models were trained on data involving English (either as a source or as a target). Nevertheless, to demonstrate that our metrics generalize well we test them on the three WMT 2019 language pairs that do not include English in either source or target. As can be seen in Table \ref{tab:not-english2019}, our results are consistent with observations in Tables \ref{tab:english-to-x2019} and \ref{tab:x-to-english2019}.  

\subsection{Robustness to High-Quality MT}
\label{sect:top-systems}

\begin{figure}[ht!]  
\centering 
\begin{tikzpicture}[scale=0.8, transform shape]
\pgfplotsset{every axis legend/.append style={
		at={(0.5,1.03)},
		anchor=south}}
\begin{axis}[
	xlabel=Top models from X to English,
	ylabel=Kendall Tau ($\tau$),
	xticklabels={ , , All, 10 , 8, 6, 4},
	legend columns=2]
\addplot[color=Unbabel7,mark=x] coordinates {
	(1,  0.327)
	(2,  0.240)
	(3,  0.198)
	(4, 0.165)
	(5 , 0.134)
};
\addplot[color=red,mark=x] coordinates {
	(1,   0.207)
	(2,  0.115)
	(3,  0.07)
	(4, 0.062)
	(5 , 0.026)
};
\addplot[color=Unbabel2,mark=x] coordinates {
	(1,  0.305)
	(2,  0.227)
	(3,  0.192)
	(4, 0.174)
	(5 , 0.150)
};
\addplot[color=Unbabel5,mark=x] coordinates {
	(1,   0.316)
	(2,  0.230)
	(3,  0.192)
	(4, 0.167)
	(5 , 0.126)
};
\addplot[color=Unbabel1,mark=x] coordinates {
	(1,  0.287)
	(2,  0.215)
	(3,  0.175)
	(4, 0.159)
	(5 , 0.143)
};
\addplot[color=Unbabel4,mark=x] coordinates {
	(1, 0.318)
	(2,  0.227)
	(3,  0.175)
	(4, 0.151)
	(5 , 0.104])
};
\legend{{\sc Comet-rank},{\sc Bleu},{\sc Comet-mqm},{\sc Bertscore},{\sc Comet-hter}, {\sc Bleurt}}
\end{axis}
\end{tikzpicture}
\begin{tikzpicture}[scale=0.8, transform shape]
\begin{axis}[
	xlabel=Top models from English to X,
	ylabel=Kendall Tau ($\tau$),
	xticklabels={ , , All, 10 , 8, 6, 4},
	]
\addplot[color=red,mark=x] coordinates {
	(1, 0.363)
	(2,  0.170)
	(3,  0.106)
	(4, 0.068)
	(5 , 0.045)
};
\addplot[color=Unbabel5,mark=x] coordinates {
	(1,  0.488)
	(2,  0.370)
	(3,  0.296)
	(4, 0.256)
	(5 , 0.226)
};
\addplot[color=Unbabel7,mark=x] coordinates {
	(1,  0.587)
	(2,  0.442)
	(3,  0.355)
	(4, 0.326)
	(5 , 0.302)
};
\addplot[color=Unbabel2,mark=x] coordinates {
	(1,  0.521)
	(2,  0.405)
	(3,  0.334)
	(4, 0.294)
	(5 , 0.260)
};
\addplot[color=Unbabel1,mark=x] coordinates {
	(1,  0.503)
	(2,  0.393)
	(3,  0.318)
	(4, 0.271)
	(5 , 0.239)
};
\end{axis}
\end{tikzpicture}
\caption{Metrics performance over all and the top (10, 8, 6, and 4) MT systems.} 
\label{fig:Top models} 
\end{figure}
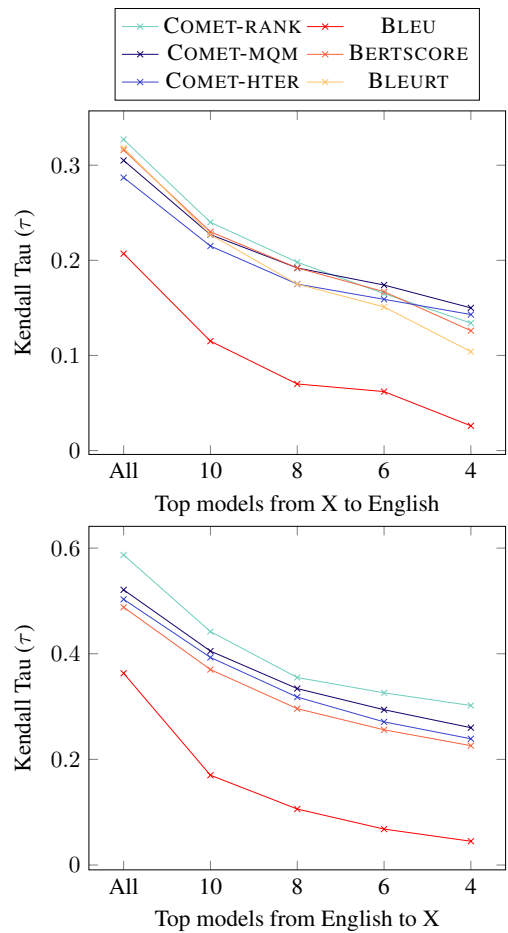

For analysis, we use the {\small DA}RR corpus from the 2019 Shared Task and evaluate on the subset of the data from the top performing MT systems for each language pair. We included language pairs for which we could retrieve data for at least ten different MT systems (i.e. all but kk-en and gu-en). We contrast against the strong recently proposed {\sc Bertscore} and {\sc Bleurt}, with {\sc Bleu} as a baseline.  Results are presented in Figure \ref{fig:Top models}. For language pairs where English is the target, our three models are either better or competitive with all others; where English is the source we note that in general our metrics exceed the performance of others. Even the MQM Estimator, trained with only 12K segments, is competitive, which highlights the power of our proposed framework.

\subsection{The Importance of the Source}

\begin{table*}[!ht]
\centering
\caption{Comparison between {\sc Comet-rank} (section \ref{ssec:ranking}) and a reference-only version thereof on WMT18 data. Both models were trained with WMT17 which means that the reference-only model is never exposed to English during training.} 
\label{tab:value-src}
\begin{tabular}{lllllllll}
\hline
\textbf{Metric}                                   & \textbf{en-cs} & \textbf{en-de} & \textbf{en-fi} & \textbf{en-tr}             & \textbf{cs-en} & \textbf{de-en} & \textbf{fi-en} & \textbf{tr-en} \\ \specialrule{1.5pt}{1pt}{1pt}
\multicolumn{1}{l|}{{\sc Comet-rank} \footnotesize ({ref. only})} & 0.660          & 0.764  & 0.630 & \multicolumn{1}{l|}{0.539} & 0.249 & 0.390  & 0.159  & 0.128 \\ 
\multicolumn{1}{l|}{{\sc Comet-rank}}   & 0.711 & 0.799 & 0.671 & \multicolumn{1}{l|}{0.563} & 0.356 & 0.542 & 0.278 & 0.260          \\ \hline
\multicolumn{1}{l|}{$\Delta \tau$}  & 0.051  & 0.035  & 0.041  & \multicolumn{1}{l|}{0.024} & \textbf{0.107} & \textbf{0.155} & \textbf{0.119} & \textbf{0.132} \\ \hline
\end{tabular}
\end{table*}

\label{sect:source-value}
To shed some light on the actual value and contribution of the source language input in our models' ability to learn accurate predictions, we trained two versions of our {\small DA}RR Ranker model: one that uses only the reference, and another that uses both reference and source. Both models were trained using the WMT 2017 corpus that only includes language pairs from English (en-de, en-cs, en-fi, en-tr). In other words, while English was never observed as a target language during training for both variants of the model, the training of the second variant includes English source embeddings.  We then tested these two model variants on the WMT 2018 corpus for these language pairs and for the reversed directions (with the exception of en-cs because cs-en does not exist for WMT 2018). The results in Table \ref{tab:value-src} clearly show that for the translation ranking architecture, including the source improves the overall correlation with human judgments. Furthermore, the inclusion of the source exposed the second variant of the model to English embeddings which is reflected in a higher $\Delta \tau$ for the language pairs with English as a target.

\section{Reproducibility}
\label{sec:reproducibility}
We will release both the code-base of the {\sc Comet} framework and the trained MT evaluation models described in this paper to the research community upon publication, along with the detailed scripts required in order to run all reported baselines.\footnote{These will be hosted at: \url{https://github.com/Unbabel/COMET}} All the models reported in this paper were trained on a single Tesla T4 (16GB) GPU. Moreover, our framework builds on top of PyTorch Lightning \cite{falcon2019pytorch}, a
lightweight PyTorch wrapper, that was created for maximal flexibility and reproducibility.

\section{Related Work}
\label{sec:literature-review}

Classic MT evaluation metrics are commonly characterized as \textbf{$n$-gram matching metrics} because, using hand-crafted features, they estimate MT quality by counting the number and fraction of $n$-grams that appear simultaneous in a candidate translation hypothesis and one or more human-references. Metrics such as {\sc Bleu} \cite{papineni-etal-2002-bleu}, {\sc Meteor} \cite{banerjee-lavie-meteor2009}, and {\sc chrF} \cite{popovic-2015-chrf} have been widely studied and improved \cite{koehn-etal-2007-moses, popovic-2017-chrf, denkowski-lavie-2011-meteor, guo-hu-2019-meteor}, but, by design, they usually fail to recognize and capture semantic similarity beyond the lexical level.

In recent years, word embeddings \cite{NIPS2013_5021, pennington-etal-2014-glove, peters-etal-2018-deep, devlin-etal-2019-bert} have emerged as a commonly used alternative to $n$-gram matching for capturing word semantics similarity. \textbf{Embedding-based metrics} like {\sc Meteor-Vector} \cite{servan-etal-2016-word2vec-vs}, {\sc Bleu2vec} \cite{tattar-fishel-2017-bleu2vec}, {\sc YiSi-1} \cite{lo-2019-yisi}, {\sc MoverScore} \cite{zhao-etal-2019-moverscore}, and {\sc Bertscore} \cite{ZhangBERTScore} create soft-alignments between reference and hypothesis in an embedding space and then compute a score that reflects the semantic similarity between those segments. However, human judgements such as DA and MQM, capture much more than just semantic similarity, resulting in a correlation upper-bound between human judgements and the scores produced by such metrics.

\textbf{Learnable metrics} \cite{shimanaka-etal-2018-ruse, mathur-etal-2019-putting, shimanaka2019machine} attempt to directly optimize the correlation with human judgments, and have recently shown promising results. {\sc Bleurt} \cite{Sellam&das-bleurt}, a learnable metric based on BERT \cite{devlin-etal-2019-bert}, claims state-of-the-art performance for the last 3 years of the WMT Metrics Shared task. Because {\sc Bleurt} builds on top of English-BERT \cite{devlin-etal-2019-bert}, it can only be used when English is the target language which limits its applicability. Also, to the best of our knowledge, all the previously proposed learnable metrics have focused on optimizing DA which, due to a scarcity of annotators, can prove inherently noisy \cite{ma-etal-2019-results}.

\textbf{Reference-less MT evaluation}, also known as Quality Estimation (QE), has historically often regressed on HTER for segment-level evaluation \cite{bojar-etal-2013-findings, bojar-etal-2014-findings, bojar-etal-2015-findings, bojar-etal-2016-findings, bojar-etal-2017-findings}. More recently, MQM has been used for document-level evaluation \cite{specia-etal-2018-findings, fonseca-etal-2019-findings}. By leveraging highly multilingual pretrained encoders such as multilingual BERT \cite{devlin-etal-2019-bert} and XLM \cite{NIPS2019_8928}, QE systems have been showing auspicious correlations with human judgements \cite{kepler-etal-2019-unbabels}. Concurrently, the OpenKiwi framework \cite{kepler-etal-2019-openkiwi} has made it easier for researchers to push the field forward and build stronger QE models. 

\section{Conclusions and Future Work}
\label{sec:conclusions}

In this paper we present {\sc Comet}, a novel neural framework for training MT evaluation models that can serve as automatic metrics and easily be adapted and optimized to different types of human judgements of MT quality. 

To showcase the effectiveness of our framework, we sought to address the challenges reported in the 2019 WMT Metrics Shared Task \cite{ma-etal-2019-results}. We trained three distinct models which achieve new state-of-the-art results for segment-level correlation with human judgments, and show promising ability to better differentiate high-performing systems. 

One of the challenges of leveraging the power of pretrained models is the burdensome weight of parameters and inference time. A primary avenue for future work on {\sc Comet} will look at the impact of more compact solutions such as DistilBERT \citep{sanh2019distilbert}.

Additionally, whilst we outline the potential importance of the source text above, we note that our {\sc Comet-rank} model weighs source and reference differently during inference but equally in its training loss function. Future work will investigate the optimality of this formulation and further examine the interdependence of the different inputs.

\section*{Acknowledgments}
We are grateful to Andr\'e Martins, Austin Matthews, Fabio Kepler, Daan Van Stigt, Miguel Vera, and the reviewers, for their valuable feedback and discussions. This work was supported in part by the P2020 Program through projects MAIA and Unbabel4EU, supervised by ANI under contract numbers 045909 and 042671, respectively.

\bibliography{anthology,emnlp2020}
\bibliographystyle{acl_natbib}

\clearpage
\appendix

\section{Appendices}

\begin{table*}[!ht]
\centering
\caption{Hyper-parameters used in our {\sc Comet} framework to train the presented models.} 
\label{tab:hyperparameters}
\begin{tabular}{lcc}
\hline
\textbf{Hyper-parameter}       & \multicolumn{1}{l}{\textbf{{\sc Comet}(Est-HTER/MQM)}} & \multicolumn{1}{l}{\textbf{{\sc Comet-rank}}} \\ \specialrule{1.5pt}{1pt}{1pt}
Encoder Model             & XLM-RoBERTa (base)                              & XLM-RoBERTa (base)                       \\
Optimizer                 & Adam (default parameters)                       & Adam (default parameters)                \\
nº frozen epochs          & 1                                               & 0                                        \\
Learning rate              & 3e-05 and 1e-05                                 & 1e-05                                    \\
Batch size                & 16                                              & 16                                       \\
Loss function             & MSE                                             & Triplet Margin ($\epsilon = 1.0$)               \\
Layer-wise dropout        & 0.1                                             & 0.1                                      \\
FP precision              & 32                                              & 32      
\\
Feed-Forward hidden units & 2304,1152                                       & --                                       \\
Feed-Forward activations  & Tanh                                            & --                                       \\
Feed-Forward dropout      & 0.1                                             & --                                       \\ \hline
\end{tabular}
\end{table*}

\label{sec:appendix}
In Table \ref{tab:hyperparameters} we list the hyper-parameters used to train our models. Before initializing these models a random seed  was set to 3 in all libraries that perform ``random'' operations (\texttt{torch}, \texttt{numpy}, \texttt{random} and \texttt{cuda}).

\begin{table*}[!ht]
\centering
\caption{Statistics for the QT21 corpus.} 
\begin{tabular}{lllll}
\hline
                      & \textbf{en-de} & \textbf{en-cs} & \textbf{en-lv} & \textbf{de-en} \\ \specialrule{1.5pt}{1pt}{1pt}
\multicolumn{1}{l|}{\textbf{Total tuples}}   & 54000 & 42000 & 35474 & 41998 \\
\multicolumn{1}{l|}{\textbf{Avg. tokens} (reference)} & 17.80 & 15.56 & 16.42 & 17.71 \\
\multicolumn{1}{l|}{\textbf{Avg. tokens} (source)} & 16.70 & 17.37 & 18.39 & 17.18 \\
\multicolumn{1}{l|}{\textbf{Avg. tokens} (MT)}  & 17.65 & 15.64 & 16.42 & 17.78 \\
\hline
\end{tabular}
\end{table*}

\begin{table*}[]
\centering
\caption{Statistics for the WMT 2017 {\footnotesize DA}RR corpus.} 
\begin{tabular}{llllll}
\hline
 & \textbf{en-cs} & \textbf{en-de} & \textbf{en-fi} & \textbf{en-lv} & \textbf{en-tr} \\ \specialrule{1.5pt}{1pt}{1pt}
\multicolumn{1}{l|}{\textbf{Total tuples}} & 32810 & 6454  & 3270  & 3456  & 247   \\
\multicolumn{1}{l|}{\textbf{Avg. tokens} (reference)} & 19.70 & 22.15 & 15.59 & 21.42 & 17.57 \\
\multicolumn{1}{l|}{\textbf{Avg. tokens} (source)} & 22.37 & 23.41 & 21.73 & 26.08 & 22.51 \\
\multicolumn{1}{l|}{\textbf{Avg. tokens} (MT)}  & 19.45 & 22.58 & 16.06 & 22.18 & 17.25 \\
\hline
\end{tabular}
\end{table*}

\begin{sidewaystable}
\centering
\caption{Statistics for the WMT 2019 {\footnotesize DA}RR into-English language pairs.} 
\begin{tabular}{llllllll}
\hline
 & \textbf{de-en} & \textbf{fi-en} & \textbf{gu-en} & \textbf{kk-en} & \textbf{lt-en} & \textbf{ru-en} & \textbf{zh-en} \\ \specialrule{1.5pt}{1pt}{1pt}
\multicolumn{1}{l|}{\textbf{Total tuples}}   & 85365 & 32179 & 20110 & 9728  & 21862 & 39852 & 31070 \\
\multicolumn{1}{l|}{\textbf{Avg. tokens} (reference)} & 20.29 & 18.55 & 17.64 & 20.36 & 26.55 & 21.74 & 42.89  \\
\multicolumn{1}{l|}{\textbf{Avg. tokens} (source)}   & 18.44 & 12.49 & 21.92 & 16.32 & 20.32 & 18.00 & 7.57   \\
\multicolumn{1}{l|}{\textbf{Avg. tokens} (MT)} & 20.22 & 17.76 & 17.02 & 19.68 & 25.25 & 21.80 & 39.70  \\
\hline
\end{tabular}
\end{sidewaystable}

\begin{sidewaystable}
\centering
\caption{Statistics for the WMT 2019 {\footnotesize DA}RR from-English and no-English language pairs.}
\begin{tabular}{llllllllllll}
\hline
 & \textbf{en-cs} & \textbf{en-de} &  \textbf{en-fi} & \textbf{en-gu} & \textbf{en-kk} & \textbf{en-lt} & \textbf{en-ru} & \textbf{en-zh} & \textbf{fr-de} & \textbf{de-cs} & \textbf{de-fr} \\ \specialrule{1.5pt}{1pt}{1pt}
\multicolumn{1}{l|}{\textbf{Total tuples}}   & 27178 & 99840 &  31820 & 11355 & 18172 & 17401 & 24334 & 18658 & 1369  & 23194 & 4862  \\
\multicolumn{1}{l|}{\textbf{Avg. tokens} (reference)} & 22.92 & 25.65 &  20.12 & 33.32 & 18.89 & 21.00 & 24.79 & 9.25  & 22.68 & 22.27 & 27.32 \\
\multicolumn{1}{l|}{\textbf{Avg. tokens} (source)}    & 24.98 & 24.97 &  25.23 & 24.32 & 23.78 & 24.46 & 24.45 & 24.39 & 28.60 & 25.22 & 21.36 \\
\multicolumn{1}{l|}{\textbf{Avg. tokens} (MT)}        & 22.60 & 24.98 &  19.69 & 32.97 & 19.92 & 20.97 & 23.37 & 6.83  & 23.36 & 21.89 & 25.68 \\
\hline
\end{tabular}
\end{sidewaystable}

\begin{sidewaystable}
\centering
\caption{MQM corpus (section \ref{sec:mqm}) statistics.} 
\begin{tabular}{lllllllllllll}
\hline
    & \textbf{en-nl} & \textbf{en-sv} & \textbf{en-ja} & \textbf{en-de} & \textbf{en-ru} & \textbf{en-es} & \textbf{en-fr} & \textbf{en-it} & \textbf{en-pt-br} & \textbf{en-tr} & \textbf{en-pt} & \textbf{en-es-latam} \\ \specialrule{1.5pt}{1pt}{1pt}
\multicolumn{1}{l|}{\textbf{Total tuples}}    & 2447  & 970   & 1590  & 2756  & 1043  & 259   & 1474  & 812   & 504      & 370   & 91    & 6           \\
\multicolumn{1}{l|}{\textbf{Avg. tokens} (reference)} & 14.10 & 14.24 & 20.32 & 13.78 & 13.37 & 10.90 & 13.75 & 13.61 & 12.48    & 7.95  & 12.18 & 10.33       \\
\multicolumn{1}{l|}{\textbf{Avg. tokens} (source)} & 14.23 & 15.31 & 13.69 & 13.76 & 13.94 & 11.23 & 12.85 & 14.22 & 12.46    & 10.36 & 13.45 & 12.33       \\
\multicolumn{1}{l|}{\textbf{Avg. tokens} (MT)}  & 13.66 & 13.91 & 17.84 & 13.41 & 13.19 & 10.88 & 13.59 & 13.02 & 12.19    & 7.99  & 12.21 & 10.17 \\
\hline
\end{tabular}
\end{sidewaystable}

\begin{sidewaystable}[]
\centering
\caption{Statistics for the WMT 2018 {\footnotesize DA}RR language pairs.}
\begin{tabular}{lllllllllllllll}
\hline
                      & \textbf{zh-en} & \textbf{en-zh} & \textbf{cs-en} & \textbf{fi-en} & \textbf{ru-en} & \textbf{tr-en} & \textbf{de-en} & \textbf{en-cs} & \textbf{en-de} & \textbf{en-et} & \textbf{en-fi} & \textbf{en-ru} & \textbf{en-tr} & \textbf{et-en} \\ \specialrule{1.5pt}{1pt}{1pt}
\multicolumn{1}{l|}{\textbf{Total tuples}}  & 33357 & 28602 & 5110  & 15648 & 10404 & 8525  & 77811 & 5413  & 19711 & 32202 & 9809  & 22181 & 1358  & 56721 \\
\multicolumn{1}{l|}{\textbf{Avg. tokens} (reference)} & 28.86 & 24.04 & 21.98 & 21.13 & 24.97 & 23.25 & 23.29 & 19.50 & 23.54 & 18.21 & 16.32 & 21.81 & 20.15 & 23.40 \\
\multicolumn{1}{l|}{\textbf{Avg. tokens} (source)} & 23.86 & 28.27 & 18.67 & 15.03 & 21.37 & 18.80 & 21.95 & 22.67 & 24.82 & 23.47 & 22.82 & 25.24 & 24.37 & 18.15 \\
\multicolumn{1}{l|}{\textbf{Avg. tokens} (MT)}  & 27.45 & 14.94 & 21.79 & 20.46 & 25.25 & 22.80 & 22.64 & 19.73 & 23.74 & 18.37 & 17.15 & 21.86 & 19.61 & 23.52 \\
\hline
\end{tabular}
\end{sidewaystable}

\newcommand{\hwplotA}{\raisebox{2pt}{\tikz{\draw[red,solid,line width=0.9pt](0,0) -- (5mm,0);}}}
\newcommand{\hwplotB}{\raisebox{2pt}{\tikz{\draw[Unbabel1,solid,line width=1.2pt](0,0) -- (5mm,0);}}}
\newcommand{\hwplotC}{\raisebox{2pt}{\tikz{\draw[Unbabel2,solid,line width=1.2pt](0,0) -- (5mm,0);}}}
\newcommand{\hwplotD}{\raisebox{2pt}{\tikz{\draw[Unbabel5,solid,line width=1.2pt](0,0) -- (5mm,0);}}}
\newcommand{\hwplotE}{\raisebox{2pt}{\tikz{\draw[Unbabel7,solid,line width=1.2pt](0,0) -- (5mm,0);}}}
\newcommand{\hwplotF}{\raisebox{2pt}{\tikz{\draw[Unbabel4,solid,line width=1.2pt](0,0) -- (5mm,0);}}}

\pgfplotsset{
	small,
}
\begin{table*}
\centering
	\begin{tikzpicture}
		\begin{axis}[
            xlabel=Top models en-cs, 
            ylabel=Kendall Tau score,
            xticklabels={ , , All, 10 , 8, 6, 4}
        ]
        \addplot[color=red,mark=x] coordinates {
        	(1, 0.364)
        	(2,  0.322)
        	(3,  0.278)
        	(4, 0.174)
        	(5 , 0.060)
        };
        
        \addplot[color=Unbabel5,mark=x] coordinates {
        	(1,  0.500)
        	(2,  0.460)
        	(3, 0.401)
        	(4, 0.305)
        	(5 , 0.202)
        };
        
        \addplot[color=Unbabel7,mark=x] coordinates {
        	(1,  0.603)
        	(2,  0.549)
        	(3,  0.503)
        	(4, 0.425)
        	(5 , 0.311)
        };
        
        \addplot[color=Unbabel2,mark=x] coordinates {
        	(1,  0.537)
        	(2,  0.493)
        	(3,  0.450)
        	(4, 0.367)
        	(5 , 0.210)
        };
        
        \addplot[color=Unbabel1,mark=x] coordinates {
        	(1,  0.524)
        	(2,  0.480)
        	(3,  0.427)
        	(4, 0.358)
        	(5 ,0.221)
        };
        \end{axis}
	\end{tikzpicture}
	\begin{tikzpicture}
	\begin{axis}[
                    xlabel=Top models en-de, 
                    xticklabels={ , , All, 10 , 8, 6, 4}
                ]
        
        \addplot[color=red,mark=x] coordinates {
        	(1, 0.248)
        	(2,  -0.039)
        	(3,  -0.099)
        	(4, -0.101)
        	(5 ,-0.152)
        };
        
        \addplot[color=Unbabel5,mark=x] coordinates {
        	(1,  0.363)
        	(2, 0.104)
        	(3, 0.046)
        	(4, 0.049)
        	(5 , 0.027)
        };
        
        \addplot[color=Unbabel7,mark=x] coordinates {
        	(1,  0.427)
        	(2,  0.142)
        	(3,  0.120)
        	(4, 0.143)
        	(5 , 0.098)
        };
        
        \addplot[color=Unbabel2,mark=x] coordinates {
        	(1,  0.398)
        	(2,  0.148)
        	(3,  0.121)
        	(4, 0.135)
        	(5 , 0.075)
        };
        
        \addplot[color=Unbabel1,mark=x] coordinates {
        	(1,  0.383)
        	(2,  0.131)
        	(3,  0.109)
        	(4, 0.113)
        	(5 ,0.110)
        };
        
        \end{axis}
	\end{tikzpicture}
	\\
	\begin{tikzpicture}
    	\begin{axis}[
                    xlabel=Top models en-fi, 
                    ylabel=Kendall Tau score,
                    xticklabels={ , , All, 10 , 8, 6, 4}
                ]
        
        \addplot[color=red,mark=x] coordinates {
        	(1, 0.395)
        	(2,  0.268)
        	(3,  0.193)
        	(4, 0.221)
        	(5 ,0.164)
        };
        
        \addplot[color=Unbabel5,mark=x] coordinates {
        	(1,  0.527)
        	(2, 0.428)
        	(3, 0.341)
        	(4, 0.391)
        	(5 , 0.355)
        };
        
        \addplot[color=Unbabel7,mark=x] coordinates {
        	(1,  0.664)
        	(2,  0.544)
        	(3,  0.412)
        	(4, 0.447)
        	(5 , 0.435)
        };
        
        \addplot[color=Unbabel2,mark=x] coordinates {
        	(1,  0.567)
        	(2,  0.470)
        	(3,  0.364)
        	(4, 0.384)
        	(5 , 0.361)
        };
        
        \addplot[color=Unbabel1,mark=x] coordinates {
        	(1,  0.560)
        	(2,  0.454)
        	(3,  0.340)
        	(4, 0.349)
        	(5 ,0.345)
        };
        
        \end{axis}
	\end{tikzpicture}%
	\begin{tikzpicture}
    	\begin{axis}[
                    xlabel=Top models en-gu, 
                    xticklabels={ , , All, 10 , 8, 6, 4}
                ]
        	
        \addplot[color=red,mark=x] coordinates {
        	(1, 0.463)
        	(2,  0.449)
        	(3,  0.438)
        	(4, 0.326)
        	(5 ,0.339)
        };
        
        \addplot[color=Unbabel5,mark=x] coordinates {
        	(1,  0.568)
        	(2, 0.556)
        	(3, 0.524)
        	(4, 0.438)
        	(5 , 0.413)
        };
        
        \addplot[color=Unbabel7,mark=x] coordinates {
        	(1,  0.611)
        	(2,  0.585)
        	(3,  0.502)
        	(4, 0.428)
        	(5 , 0.316)
        };
        
        \addplot[color=Unbabel2,mark=x] coordinates {
        	(1,  0.564)
        	(2,  0.542)
        	(3,  0.488)
        	(4, 00.405)
        	(5 , 0.363)
        };
        
        \addplot[color=Unbabel1,mark=x] coordinates {
        	(1,  0.552)
        	(2,  0.528)
        	(3,  0.468)
        	(4, 0.363)
        	(5 ,0.253)
        };
        \end{axis}
	\end{tikzpicture}%
	\\
	\begin{tikzpicture}
	    \begin{axis}[
            xlabel=Top models en-kk, 
            ylabel=Kendall Tau score,
            xticklabels={ , , All, 10 , 8, 6, 4}
        ]
        
        \addplot[color=red,mark=x] coordinates {
        	(1, 0.363)
        	(2,  0.175)
        	(3,  0.083)
        	(4, 0.036)
        	(5 ,-0.039)
        };
        
        \addplot[color=Unbabel5,mark=x] coordinates {
        	(1,  0.540)
        	(2, 0.376)
        	(3, 0.260)
        	(4, 0.207)
        	(5 , 0.115)
        };
        
        \addplot[color=Unbabel7,mark=x] coordinates {
        	(1,  0.693)
        	(2,  0.374)
        	(3,  0.291)
        	(4, 0.237)
        	(5 , 0.182)
        };
        
        \addplot[color=Unbabel2,mark=x] coordinates {
        	(1,  0.534)
        	(2,  0.358)
        	(3,  0.263)
        	(4, 0.231)
        	(5 , 0.195)
        };
        
        \addplot[color=Unbabel1,mark=x] coordinates {
        	(1,  0.508)
        	(2,  0.322)
        	(3,  0.234)
        	(4, 0.192)
        	(5 ,0.182)
        };
        \end{axis}  
	\end{tikzpicture}
	\begin{tikzpicture}
	    \begin{axis}[
            xlabel=Top models en-lt, 
            xticklabels={ , , All, 10 , 8, 6, 4}
        ]
        
        \addplot[color=red,mark=x] coordinates {
        	(1, 0.333)
        	(2,  0.249)
        	(3,  0.12)
        	(4, 0.062)
        	(5 ,0.023)
        };
        
        \addplot[color=Unbabel5,mark=x] coordinates {
        	(1,  0.464)
        	(2, 0.369)
        	(3, 0.235)
        	(4, 0.175)
        	(5 , 0.131)
        };
        
        \addplot[color=Unbabel7,mark=x] coordinates {
        	(1,  0.665)
        	(2,  0.550)
        	(3,  0.358)
        	(4, 0.300)
        	(5 , 0.305)
        };
        
        \addplot[color=Unbabel2,mark=x] coordinates {
        	(1,  0.574)
        	(2,  0.473)
        	(3,  0.327)
        	(4, 0.290)
        	(5 , 0.261)
        };
        
        \addplot[color=Unbabel1,mark=x] coordinates {
        	(1,  0.577)
        	(2,  0.478)
        	(3,  0.327)
        	(4, 0.251)
        	(5 ,0.226)
        };
        \end{axis}
        \end{tikzpicture}
    \\
    \begin{tikzpicture}
        \begin{axis}[
            xlabel=Top models en-ru, 
            ylabel=Kendall Tau score,
            xticklabels={ , , All, 10 , 8, 6, 4}
        ]
        
        \addplot[color=red,mark=x] coordinates {
        	(1, 0.469)
        	(2,  0.289)
        	(3,  0.170)
        	(4, 0.112)
        	(5 ,0.185)
        };
        
        \addplot[color=Unbabel5,mark=x] coordinates {
        	(1,  0.585)
        	(2, 0.414)
        	(3, 0.317)
        	(4, 0.268)
        	(5 , 0.331)
        };
        
        \addplot[color=Unbabel7,mark=x] coordinates {
        	(1,  0.580)
        	(2,  0.468)
        	(3,  0.391)
        	(4, 0.388)
        	(5 , 0.450)
        };
        
        \addplot[color=Unbabel2,mark=x] coordinates {
        	(1,  0.615)
        	(2,  0.488)
        	(3,  0.416)
        	(4, 0.320)
        	(5 , 0.366)
        };
        
        \addplot[color=Unbabel1,mark=x] coordinates {
        	(1,  0.539)
        	(2,  0.480)
        	(3,  0.395)
        	(4, 0.332)
        	(5 ,0.358)
        };
        \end{axis}
    \end{tikzpicture}
    \begin{tikzpicture}
        \begin{axis}[
            xlabel=Top models en-zh, 
            xticklabels={ , , All, 10 , 8, 6, 4}
            legend style={font=\tiny},legend columns=2]
        ]
        	
        \addplot[color=red,mark=x] coordinates {
        	(1, 0.235)
        	(2, 0.128)
        	(3,  0.146)
        	(4, 0.135)
        	(5 ,0.165)
        };
        
        \addplot[color=Unbabel5,mark=x] coordinates {
        	(1,  0.356)
        	(2, 0.255)
        	(3, 0.244)
        	(4, 0.218)
        	(5 , 0.236)
        };
        
        \addplot[color=Unbabel7,mark=x] coordinates {
        	(1,  0.449)
        	(2,  0.324)
        	(3,  0.267)
        	(4, 0.243)
        	(5 , 0.319)
        };
        
        \addplot[color=Unbabel2,mark=x] coordinates {
        	(1,  0.378)
        	(2,  0.265)
        	(3,  0.241)
        	(4, 0.219)
        	(5 , 0.248)
        };
        
        \addplot[color=Unbabel1,mark=x] coordinates {
        	(1,  0.380)
        	(2,  0.274)
        	(3,  0.240)
        	(4, 0.214)
        	(5 ,0.223)
        };
        \end{axis}
        
    \end{tikzpicture}
\caption{Metrics performance over all and the top (10,8, 6, and 4) MT systems for all from-English language pairs. The color scheme is as follows: {\hwplotE} {\sc Comet-rank}, {\hwplotB} {\sc Comet-hter}, {\hwplotC} {\sc Comet-mqm}, {\hwplotA} {\sc Bleu}, {\hwplotF} {\sc Bertscore}}
\end{table*}%

\pgfplotsset{
	small,
}
\begin{table*}
\centering
    \begin{tabular}{ c c } 
	\begin{tikzpicture}
        \begin{axis}[
            xlabel=Top models de-en, 
            ylabel=Kendall Tau score,
            xticklabels={ , , All, 10 , 8, 6, 4}
        ]
        
        \addplot[color=red,mark=x] coordinates {
        	(1, 0.053)
        	(2,  -0.049)
        	(3,  -0.06)
        	(4, -0.049)
        	(5 ,-0.080)
        };
        
        \addplot[color=Unbabel5,mark=x] coordinates {
        	(1,  0.190)
        	(2,  0.094)
        	(3,  0.093)
        	(4, 0.106)
        	(5 , 0.095)
        };
        
        \addplot[color=Unbabel7,mark=x] coordinates {
        	(1,  0.202)
        	(2,  0.118)
        	(3,  0.108)
        	(4, 0.130)
        	(5 , 0.131)
        };

        \addplot[color=Unbabel2,mark=x] coordinates {
        	(1,  0.207)
        	(2,  0.137)
        	(3,  0.135)
        	(4, 0.152)
        	(5 , 0.150)
        };

        \addplot[color=Unbabel1,mark=x] coordinates {
        	(1,  0.185)
        	(2,  0.128)
        	(3,  0.122)
        	(4, 0.141)
        	(5 , 0.140)
        };
        
        \addplot[color=Unbabel4,mark=x] coordinates {
        	(1, 0.174)
        	(2,  0.053)
        	(3,  0.030)
        	(4, 0.055)
        	(5 , 0.049])
        };
	    \end{axis}
	\end{tikzpicture} &
	\begin{tikzpicture}
	    \begin{axis}[
            xlabel=Top models fi-en, 
            xticklabels={ , , All, 10 , 8, 6, 4}
        ]
        	
        \addplot[color=red,mark=x] coordinates {
        	(1, 0.236)
        	(2,  0.182)
        	(3,  0.167)
        	(4, 0.144)
        	(5 , 0.104)
        };
        
        \addplot[color=Unbabel5,mark=x] coordinates {
        	(1,  0.354)
        	(2,  0.314)
        	(3,  0.282)
        	(4, 0.242)
        	(5 , 0.179)
        };
        
        \addplot[color=Unbabel7,mark=x] coordinates {
        	(1,  0.399)
        	(2,  0.326)
        	(3,  0.292)
        	(4, 0.232)
        	(5 , 0.143)
        };
         
        \addplot[color=Unbabel2,mark=x] coordinates {
        	(1,  0.343)
        	(2,  0.298)
        	(3,  0.269)
        	(4, 0.226)
        	(5 , 0.165)
        };

        \addplot[color=Unbabel1,mark=x] coordinates {
        	(1,  0.333)
        	(2,  0.287)
        	(3,  0.244)
        	(4, 0.208)
        	(5 , 0.157)
        };
        
        \addplot[color=Unbabel4,mark=x] coordinates {
        	(1, 0.374)
        	(2,  0.335)
        	(3,  0.299)
        	(4, 0.262)
        	(5 , 0.216])
        };
        \end{axis}
    \end{tikzpicture} \\
    \begin{tikzpicture}
        \begin{axis}[
            xlabel=Top models lt-en, 
            ylabel=Kendall Tau score,
            xticklabels={ , , All, 10 , 8, 6, 4}
        ]
        	
        \addplot[color=red,mark=x] coordinates {
        	(1,0.249)
        	(2, 0.226)
        	(3, 0.1130)
        	(4, .078)
        	(5 ,0.048)
        };
        
        \addplot[color=Unbabel5,mark=x] coordinates {
        	(1,  0.381)
        	(2, 0.348)
        	(3,  0.256)
        	(4, 0.175)
        	(5 , 0.162)
        };
        
        \addplot[color=Unbabel7,mark=x] coordinates {
        	(1,  0.407)
        	(2,  0.360)
        	(3,  0.260)
        	(4, 0.192)
        	(5 , 0.142)
        };
        \addplot[color=Unbabel2,mark=x] coordinates {
        	(1,  0.368)
        	(2,  0.330)
        	(3,  0.229)
        	(4, 0.185)
        	(5 , 0.172)
        };
        \addplot[color=Unbabel1,mark=x] coordinates {
        	(1,  0.364)
        	(2,  0.326)
        	(3,  0.223)
        	(4, 0.158)
        	(5 , 0.146)
        };
        
        \addplot[color=Unbabel4,mark=x] coordinates {
        	(1, 0.388)
        	(2,  0.356)
        	(3,  0.266)
        	(4, 0.181)
        	(5 , 0.154])
        };
        \end{axis}
    \end{tikzpicture} &
    \begin{tikzpicture}
        \begin{axis}[
            xlabel=Top models ru-en, 
            xticklabels={ , , All, 10 , 8, 6, 4},
            ymin=0
        ]
        
        \addplot[color=red,mark=x] coordinates {
        	(1, 0.177)
        	(2,  0.071)
        	(3, 0.055)
        	(4, 0.048)
        	(5 , 0.065)
        };
        
        \addplot[color=Unbabel5,mark=x] coordinates {
        	(1,  0.221)
        	(2,  0.168)
        	(3,  0.146)
        	(4, 0.143)
        	(5 , 0.137)
        };
        
        \addplot[color=Unbabel7,mark=x] coordinates {
        	(1,  0.180)
        	(2,  0.152)
        	(3,  0.125)
        	(4, 0.105)
        	(5 ,0.140)
        };
         
        \addplot[color=Unbabel2,mark=x] coordinates {
        	(1,  0.187)
        	(2,  0.138)
        	(3,  0.117)
        	(4, 0.105)
        	(5 , 0.107)
        };
         
        \addplot[color=Unbabel1,mark=x] coordinates {
        	(1,  0.163)
        	(2,  0.132)
        	(3,  0.114)
        	(4, 0.098)
        	(5 , 0.119)
        };
        
        \addplot[color=Unbabel4,mark=x] coordinates {
        	(1, 0.220)
        	(2,  0.171)
        	(3,  0.147)
        	(4, 0.134)
        	(5 , 0.167])
        };
        \end{axis}
    \end{tikzpicture}
    \\
    \begin{tikzpicture}
        \begin{axis}[
        	xlabel=Top models zh-en,
        	ylabel=Kendall Tau score,
        	xticklabels={ , , All, 10 , 8, 6, 4},
        ]

        \addplot[color=Unbabel7,mark=x] coordinates {
        	(1,  0.445)
        	(2,  0.245)
        	(3,  0.205)
        	(4, 0.167)
        	(5 , 0.114)
        };
        	
        \addplot[color=red,mark=x] coordinates {
        	(1,0.321)
        	(2, 0.144)
        	(3, 0.064)
        	(4,0.088)
        	(5 ,-0.005)
        };
        \addplot[color=Unbabel2,mark=x] coordinates {
        	(1,  0.422)
        	(2,  0.231)
        	(3,  0.210)
        	(4, 0.204)
        	(5 , 0.154)
        };
        
        \addplot[color=Unbabel5,mark=x] coordinates {
        	(1,  0.432)
        	(2, 0.228)
        	(3,  0.182)
        	(4, 0.169)
        	(5 , 0.055)
        };
        
        \addplot[color=Unbabel1,mark=x] coordinates {
        	(1,  0.391)
        	(2,  0.201)
        	(3,  0.172)
        	(4, 0.190)
        	(5 , 0.154)
        };
        
        \addplot[color=Unbabel4,mark=x] coordinates {
        	(1, 0.436)
        	(2,  0.221)
        	(3,  0.131)
        	(4, 0.124)
        	(5 ,-0.065])
        };
        \end{axis}
    \end{tikzpicture}
    \end{tabular}
\caption{Metrics performance over all and the top (10,8, 6, and 4) MT systems for all into-English language pairs. The color scheme is as follows: {\hwplotE} {\sc Comet-rank}, {\hwplotB} {\sc Comet-hter}, {\hwplotC} {\sc Comet-mqm}, {\hwplotA} {\sc Bleu}, {\hwplotF} {\sc Bertscore} , {\hwplotD} {\sc Bleurt}}
\end{table*}%

\end{document}